\documentclass[10pt,twocolumn,letterpaper]{article}

\usepackage[pagenumbers]{cvpr} % To force page numbers, e.g. for an arXiv version

\usepackage{graphicx}
\usepackage{amsmath}
\usepackage{amssymb}
\usepackage{booktabs}
\usepackage{paralist}
\usepackage{float}

\usepackage[pagebackref,breaklinks,colorlinks]{hyperref}

\usepackage[capitalize]{cleveref}
\crefname{section}{Sec.}{Secs.}
\Crefname{section}{Section}{Sections}
\Crefname{table}{Table}{Tables}
\crefname{table}{Tab.}{Tabs.}

\def\cvprPaperID{*****} % *** Enter the CVPR Paper ID here
\def\confName{AI535}
\def\confYear{S2022}

\begin{document}

%%%%%%%%% TITLE - PLEASE UPDATE
\title{ Contrastive Learning for Object Detection}

\author{Rishab Balasubramanian \hspace{40pt} Kunal Rathore\\
Oregon State University\\
{\tt\small \{balasuri, rathorek\}@oregonstate.edu}
}
\maketitle

\section{Introduction} 

\subsection{Problem Statement}
In this project we will be working on Object Detection using Contrastive Learning. The goal of the project is to implement and evaluate the success of contrastive learning paradigm for learning better feature representations, and use these for object detection. \\
%
% What's contrastive learning?
%Mentioning about OOD?
% Motivations?
%   - state of art Contrastive learning: better to site some papers here as well.

Contrastive Learning follows from traditional triplet-loss, where similarity between an ``anchor" and ``positive" is maximized, and the similarity between ``anchor" and ``negative" is minimized. Contrastive learning is commonly used as a method of self-supervised learning with the ``anchor" and ``positive" being two random augmentations of a given input image, and the ``negative" is the set of all other images. This has been shown to outperform traditional approaches such as the triplet loss and N-pair loss in \cite{chen2020simple}. However, the requirement of large batch sizes and memory banks has made it difficult and slow to train (\cite{chen2020simple}, \cite{he2020momentum}, \cite{chen2020improved}). This motivated the rise of Supervised Contrasative approaches that overcome these problems by using annotated data \cite{khosla2020supervised}. However there is no explicit emphasis on learning good representation, but rather the idea is to cluster points into regions such that they are separable in the higher dimensional parameter space. The authors in \cite{hoffmann2022ranking} make an attempt at enforcing better representation learning into the contrastive learning framework by clustering classes together based on their similarity to each other.\\

Inspired by this approach, we look to rank classes based on their similarity, and observe the impact of human bias (in the form of ranking) on the learned representations. We feel this is an important question to address, as learning good feature embeddings has been a long sought after problem in computer vision. This would also be important for similar domains such as OOD detection, image matching/retrieval, and other tasks which require a good representation of the images. Code available at \texttt{\url{https://github.com/rishabbala/Contrastive_Learning_For_Object_Detection}}\\

\subsection{Scope and Challenges}
We work in a supervised setting with labels for the bounding boxes of objects and their corresponding class. We work with VOC 2007, a sufficiently large dataset of annotated images belonging to 20 different classes. The main challenge in our work would be the training time for multiple experiments. Due to the longer training time, and batch size of contrastive learning, we would need access to GPUs for our experiments. The class similarity rankings, which are a user input are also a challenge as they would require manual tuning and multiple experiments to identify useful ranking order.\\

\section{Approach}

\begin{figure*}
    \centering
    \subfloat{{\includegraphics[width=0.95\textwidth, height=2.5cm]{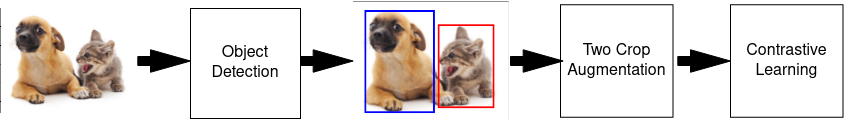} }}\\
\caption{Our Approach}
\label{fig:approach}
\end{figure*}

Fig \ref{fig:approach} shows the approach we follow for training our model. Given an input image, we pass it through the Object Detection module to predict bounding boxes (Sec \ref{sub:OD}). Once the bounding boxes are predicted, we perform a Two-Crop transformation on each object in the image (Sec \ref{sub:twocrop}), and pass it through our Contrastive Learning framework (Sec \ref{sub:OC}). We divide the process into three main stages
\begin{itemize}
    \item Object Detection
    \item Two Crop Augmentations
    \item Object Classification
\end{itemize}

\subsection{Object Detection}
\label{sub:OD}
\begin{figure}
    \centering
    \subfloat{{\includegraphics[width=0.5\textwidth]{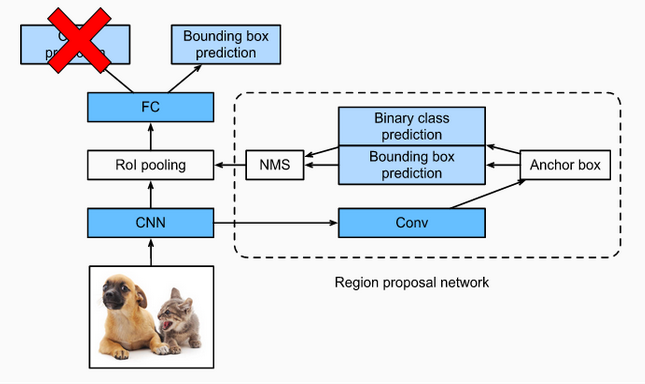} }}\\
\caption{Faster RCNN}
\label{fig:Faster_RCNN}
\end{figure}

In this work we use Faster RCNN \cite{ren2015faster} for object detection. Faster RCNN has two main components, a region proposal network (RPN), and a classification network. We remove the classification network and retain only the proposed bounding boxes by the RPN. Fig \ref{fig:Faster_RCNN} (adapted from 
\href{https://d2l.ai/}{d2l}) shows the Faster RCNN pipeline used in our work. Given the location of the bounding boxes, the image is cropped at these locations and a two-crop transformation is performed.\\

\subsection{Two Crop Transformation}
\label{sub:twocrop}
\begin{figure*}
    \centering
    \subfloat{{\includegraphics[width=0.95\textwidth, height=6cm]{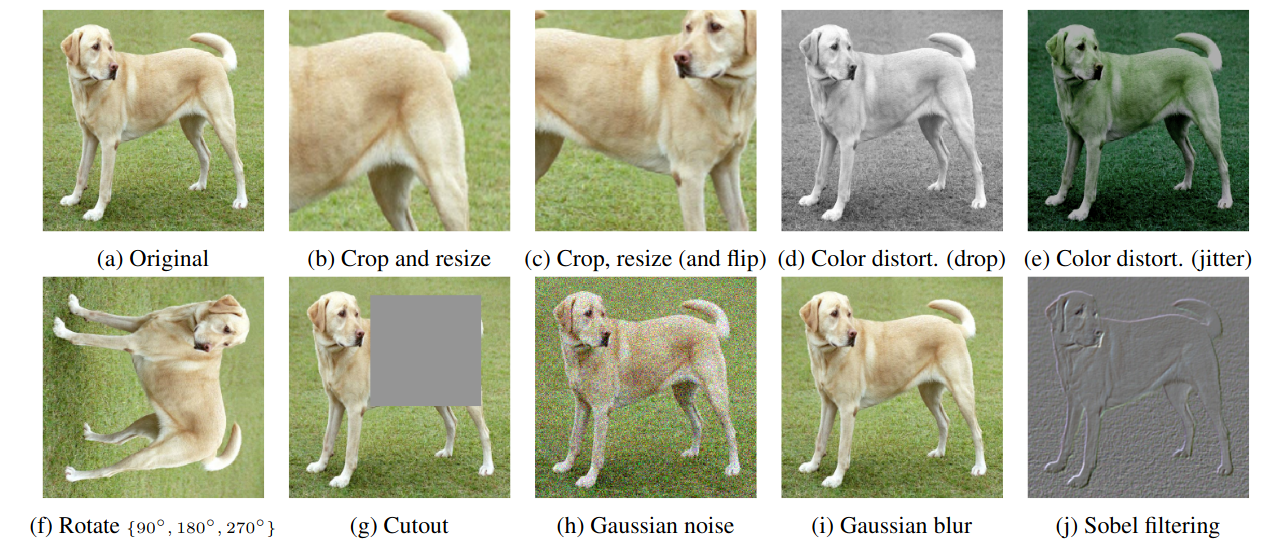} }}\\
\caption{Proposed transformations}
\label{fig:transformations}
\end{figure*}

The cropped images at the location of the bounding boxes are stacked together into a batch of images. We first normalize the images using the mean and standard deviation of the dataset. Then we follow the standard transformations for Contrastive Learning proposed in \cite{khosla2020supervised}, as shown in Fig  \ref{fig:transformations}. Of these we do not use the cutout, blur, and sobel filter augmentations proposed. We perform two random combination using a subset of these methods to produce two augmentations of the image. The first is the ``anchor", and the second falls in the ``positive" class.\\

\subsection{Object Classification}
\label{sub:OC}
\begin{figure*}
    \centering
    \subfloat{{\includegraphics[width=0.95\textwidth]{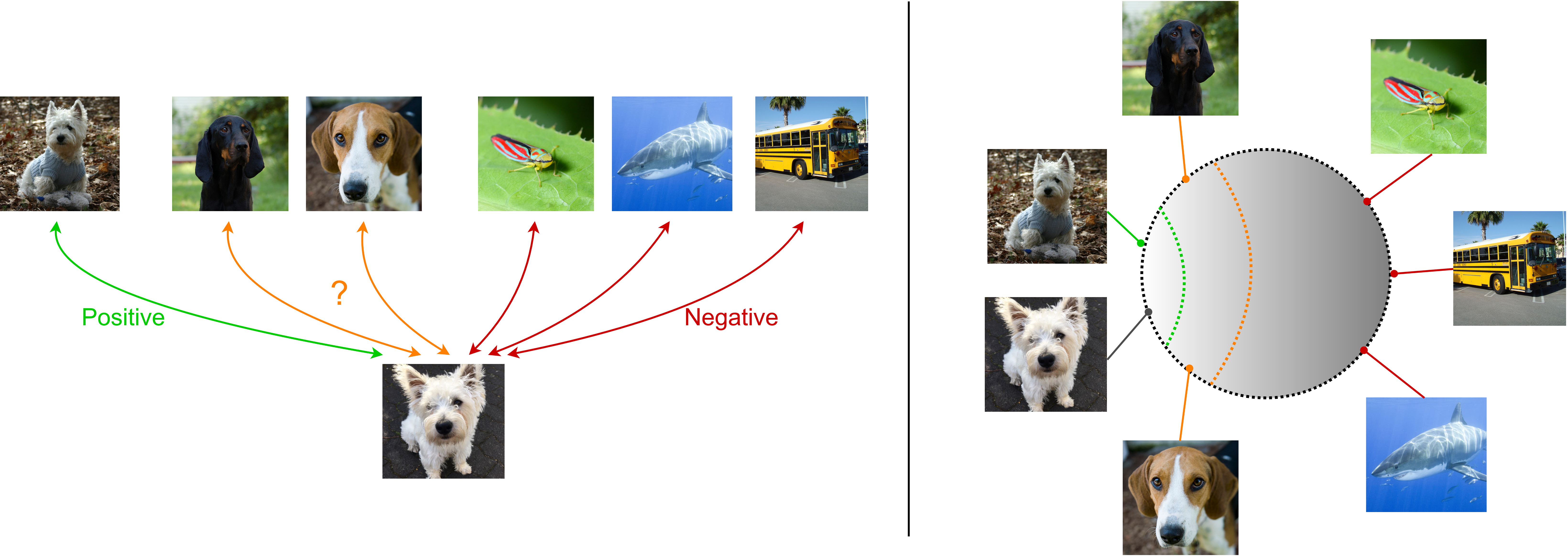} }}\\
\caption{Traditional Contrastive Learning approaches are binary (left), where there is a single ``anchor", and a single ``positive" image/class. We use a ranking system to improve learned features (right)}
\label{fig:CL}
\end{figure*}

Fig \ref{fig:CL} (adapted from \cite{hoffmann2022ranking}) shows the Contrastive Learning approach we use. We follow \cite{hoffmann2022ranking}, and use a ranked supervised contrastive learning method, where the ranking is user defined. This is different from traditional approaches which use a single ``positive" image/class.\\

For each anchor (query) image $q$, we rank a number of similar classes as $\mathcal{P}_1 \cdots \mathcal{P}_r$, where $r$ denotes the number of positive classes in our ranking. We also define a negative class as $\mathcal{N}$. Let $h(q,x)$ be the cosine similarity between the query and any other image $x$. Then, we can define our objective as enforcing :

\begin{equation}
h(q, \mathcal{P}_1) > h(q, \mathcal{P}_2) > \cdots h(q, \mathcal{P}_r) > h(q, \mathcal{N})
\end{equation}

We do this by defining a loss $L = \sum_{i=1}^{r} l_{i}$ where

\begin{equation}
    \mathit{l_{i}} = -\log \frac{\sum_{p \in \mathcal{P}_{i}} \exp(\frac{h(q,p)}{\tau_{i}})}{\sum_{p \in \cup_{j \geq i} \mathcal{P}_{i}} \exp(\frac{h(q,p)}{\tau_{i}}) + \sum_{n \in \mathcal{N} \exp(\frac{h(q,n)}{\tau_{i}})}}    
\end{equation}

This can be thought of as recursively computing the loss ($L$), when considering the current highest ranked class ($i$) as ``positive" and all other classes as negative. After computing the loss, the current highest ranked class ($i$) is removed, and the loss is computed again for class $i+1$. To ensure good separation, we set $\tau_{i+1} > \tau_{i}$, following the empirical studies provided in \cite{hoffmann2022ranking}.\\

Opposed to \cite{hoffmann2022ranking}, we rank classes instead of clustering them into groups. The difference between them is that in our ranking, any class could be ranked similar to any other class, with a user defined score. However, in clustering as done in \cite{hoffmann2022ranking}, only the classes within the same cluster are considered similar. For example, \cite{hoffmann2022ranking} puts the classes ``aeroplane" and ``ship" together as ``vehicles". However, from human knowledge, we know that an ``aeroplane" is also (probably more) similar to a ``bird" than a ``ship". Hence in our method, we create the ranking for the class ``aeroplane" as \{``bird", ``ship", ... \} with decreasing order of similarity from left to right.\\

\section{Evaluation}
\subsection{Implementation Details}
\begin{itemize}
    \item We use the detectron2 library \cite{wu2019detectron2} open sourced by Facebook for the bounding box predictions. We use a ResNet 50 FPN backbone for object detection
    \item We build upon the code provided in \cite{hoffmann2022ranking} to incorporate our ranking and experiments.
    \item We used ResNet backbone 50 for all our Contrastive Learning experiments
    \item We run our experiments with a batch size of 32 with the VOC2007 dataset
    \item We trained our model for 500 epochs
    \item We use cosine similarity, with a learning rate of 0.5, and a learning decay rate of 0.1.
    \item We set the temperature in the loss $\tau \in [0.1, 0.6]$
    \item The experiments are conducted with the number of positively ranked classes $r \in \{1, 3, 5\}$
\end{itemize}

\subsection{Dataset}
We evaluate on the VOC2007 dataset, which has 20 classes, 9963 images and 24640 objects (bounding boxes). This is split into 5011 images in the training/val set and 4952 images in the test set, with similar number of objects between the two. The dataset provides the images, ground truth bounding box annotations, and the corresponding class for each bounding box. \\
% We also train and evaluate on CIFAR10 which has 50000 training images and 10000 testing images belonging to 10 classes.\\

\subsection{Metrics And Comparison}
We evaluate the mAP for the object detection stage and evaluate classification accuracy for the Contrastive Learning model. We compare the accuracy with SupCL (\cite{khosla2020supervised} where there is no ranking), RINCE (\cite{hoffmann2022ranking}, where similar classes are clustered together), and SoftMax (common discriminative approach of training a ResNet 50 with a SoftMax loss). \textbf{Extra Credit:} We also test our model for detecting OOD classes. In this case we plot the ROC curve of True positive Rate vs False Positive Rate and compare the Area Under the Curve.\\

\begin{table*}[t]
\resizebox{\textwidth}{!}{%
\begin{tabular}{ |p{8cm}|p{2cm}|p{2cm}|p{2cm}|  }
 \hline
 Model & AP & AP50 & AP75 \\
 \hline
Faster RCNN (trained on VOC2007 train+val) ResNet50 FPN & 45.254 & 72.746 & 49.338\\
\hline
\end{tabular}}
\caption{Object Detection scores}
\label{tab:map}
\end{table*}

\subsection{Results \& Evaluations}

\begin{figure*}[h!]
    \centering
    \subfloat{{\includegraphics[width=0.3\textwidth, height=5cm]{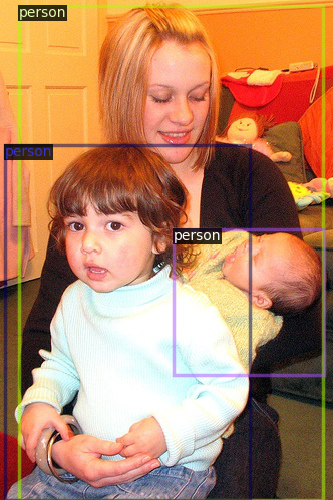} }}
    \subfloat{{\includegraphics[width=0.3\textwidth, height=5cm]{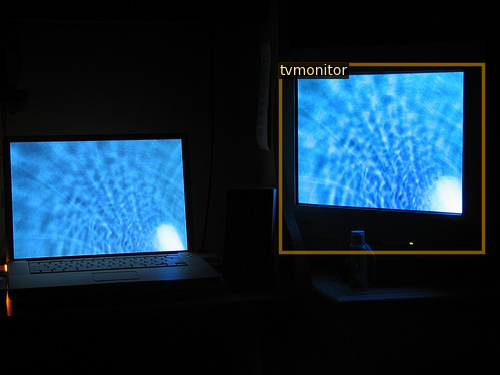} }}
    \subfloat{{\includegraphics[width=0.3\textwidth, height=5cm]{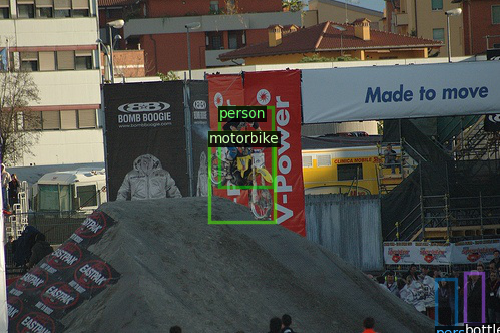} }}\\
    \subfloat{{\includegraphics[width=0.3\textwidth, height=5cm]{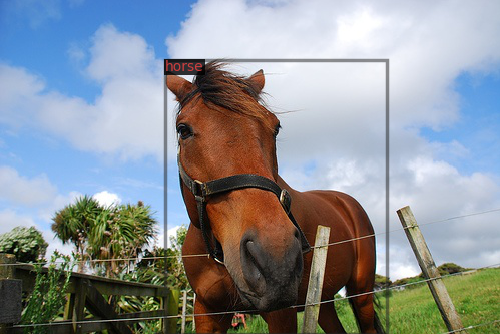} }}
    \subfloat{{\includegraphics[width=0.3\textwidth, height=5cm]{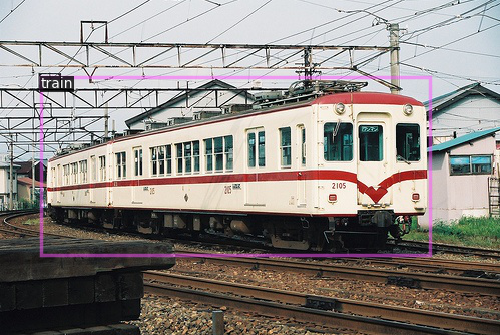} }}
    \subfloat{{\includegraphics[width=0.3\textwidth, height=5cm]{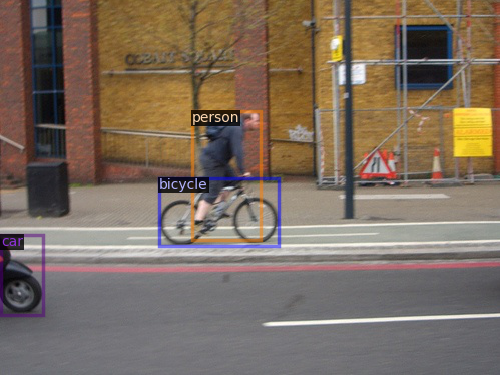} }}\\
\caption{Results}
\label{fig:results}
\end{figure*}

We first evaluate the efficiency of the Faster-RCNN model in predicting bounding boxes. To do so, we compute the AP scores using a pre-trained model. Table \ref{tab:map} shows the AP, AP50, and AP75 scores of the Object detection module we used. This is lower than the values reported in the Faster RCNN paper, and also lower than more modern approaches. Since the objective was not only object detection, we did not try different detection models.\\ 

Table \ref{table:VOC} shows the classification accuracy on the VOC2007 dataset, and Fig \ref{fig:results} shows the results from our model. We observe that again our accuracy is comparable to RINCE (\cite{hoffmann2022ranking}) and SupCL (\cite{khosla2020supervised}), while the discriminative classification receives a much lower score. However, we observed that our scores are slightly lesser than SupCL and RINCE. Since, all other parameters were similar during testing, the two factors that affects these results the most is ranking, and the user-tuned class-similarity scores. Due to the diverse nature of classes in the dataset, it resulted in sparser ranking which affects the performance of this approach.\\

\begin{table}[h]
  \centering
  \begin{tabular}{@{}lc@{}}
    \toprule
    Method & Classification Accuracy \\
    \midrule
    SupCL ($r$=1) & 0.6499\\
    Ours($r$=3) & 0.6068\\
    RINCE($r$=5) & 0.6368\\
    SoftMax & 0.5829\\
    \bottomrule
  \end{tabular}
   \caption{Classification Accuracy on VOC2007}
 \label{table:VOC}
\end{table}

\begin{figure*}[h!]
    \centering
    \subfloat[\centering baseline]{{\includegraphics[width=0.3\textwidth, height=5cm]{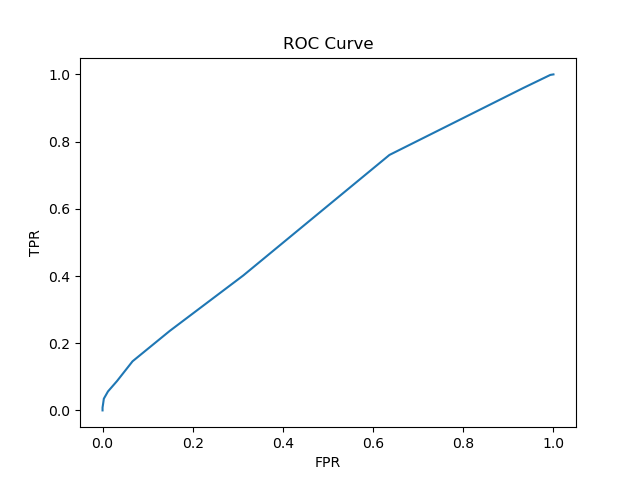} }}
    \subfloat[\centering 1 positive class ($r=1$)]{{\includegraphics[width=0.3\textwidth, height=5cm]{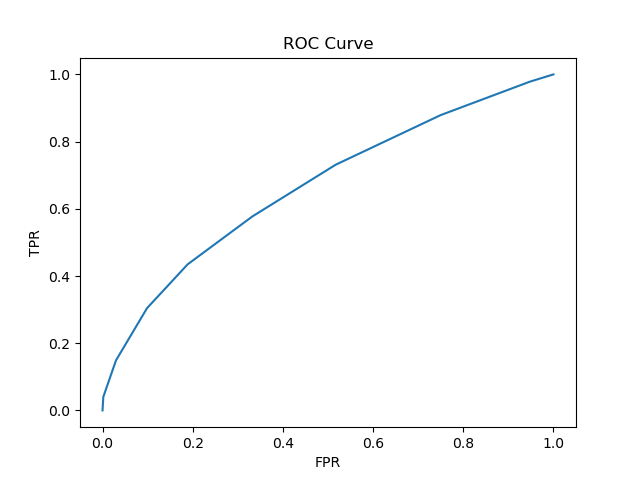} }}
    \subfloat[\centering 5 positive classes ($r=5$)]{{\includegraphics[width=0.3\textwidth, height=5cm]{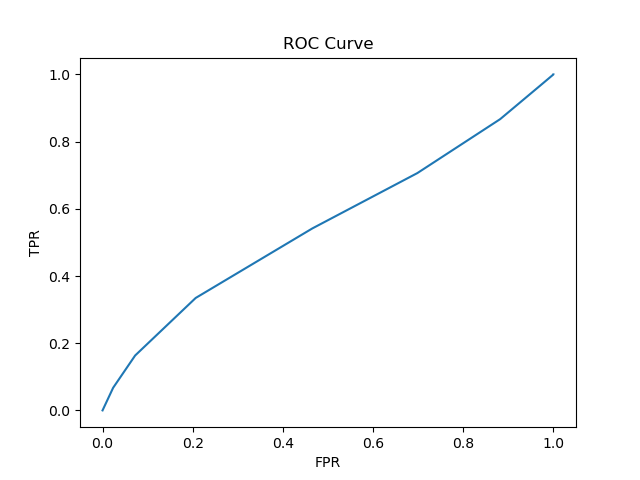} }}\\
\caption{ROC plots for VOC}
\label{fig:rocVOC}
\end{figure*}

\subsection{Extra Credit: Out of Distribution Detection}
We finally evaluate our model's performance for OOD object detection. We evaluate on the VOC2007 dataset, with 2 classes withheld and show the ROC curve in Fig \ref{fig:rocVOC} and the AUROC in Table \ref{table:VOCOOD}. We see that our model does not perform better than the baselines. This shows that our method enforces good representation learning when human input is given, but the representations for new classes are poor. We can conclude that given an unobserved object class, our model pushes it close to one of the known classes thus resulting in poor results.

% \begin{table}[h]
%   \centering
%   \begin{tabular}{@{}lc@{}}
%     \toprule
%     Method & AUROC \\
%     \midrule
%     SupCL ($r$=1) & 0.7426\\
%     Ours($r$=3) & 0.4911\\
%     SoftMax & 0.611\\
%      \hline
%     \end{tabular}
%      \caption{AUROC on CIFAR10 with 2 classes witheld}
%      \label{table:CIFAR10OOD}
% \end{table}

\begin{table}[h]
  \centering
  \begin{tabular}{@{}lc@{}}
    \toprule
    Method & AUROC \\
    \midrule
    SupCL ($r$=1) & 0.6679\\
    Ours($r$=5) & 0.5532\\
    SoftMax & 0.5621\\
 \hline
\end{tabular}
 \caption{AUROC on VOC2007 with 2 classes withheld}
 \label{table:VOCOOD}
\end{table}

% \begin{figure*}[h!]
%     \centering
%     \subfloat[\centering 1 positive class]{{\includegraphics[width=0.3\textwidth, height=5cm]{ROC_Cifar10_baseline.png} }}
%     \subfloat[\centering 5 positive class]{{\includegraphics[width=0.3\textwidth, height=5cm]{ROC_cifar10_ranking_depth_1.png} }}
%     \subfloat[\centering baseline]{{\includegraphics[width=0.3\textwidth, height=5cm]{ROC_cifar10_ranking_depth_5.png} }}\\
% \caption{ROC plots for CIFAR10}
% \label{fig:rocCIFAR10}
% \end{figure*}

\subsection{Runtimes \& Hardware}
We train all our models on the HPC cluster using a Tesla V100 GPU. For training we use a batch size of 32, and observe that it takes around 1.5-2 minutes per epoch During testing we observe that we take 2 minutes to to generate results over the validation set. Evaluation of the Average Precision scores for Faster RCNN takes about 5 minutes.\\

\subsection{Individual Contributions}
Kunal worked on the object detection pipeline, and its evaluation. Rishab worked on setting up the Contrastive Learning experiments and training them. We worked together for OOD Detection. We filled up our respective portions in the report, and made changes together.\\

%%%%%%%%% REFERENCES
{\small
\bibliographystyle{ieee_fullname}
\bibliography{egbib}
}
\end{document}